\documentclass{article}
\usepackage[whole]{bxcjkjatype} 
\usepackage[unicode]{hyperref}

\usepackage{newtxtext,newtxmath}
\usepackage{url}
\usepackage{pcsjimps-j}
\usepackage{graphicx}
\usepackage{hyperref}

\begin{document}
\pagestyle{empty}

\twocolumn[\centering
\JTitle{自動インフラ点検のためのメータ検出手法の比較検討}
\ETitle{A Comparative Study of Meter Detection Methods for Automated Infrastructure Inspection}

\JEAuthor{大坪悠介}{Yusuke Ohtsubo}{35mm}
\JEAuthor{佐藤拓杜}{Takuto Sato}{35mm}
\JEAuthor{佐川浩彦}{Hirohiko Sagawa}{35mm}\\
\JEAffiliation{日立製作所 中央研究所}{ Hitachi, Ltd. Central Research Laboratory}{70mm}
\\

\Abstract{
位置誤差を含む自律点検ロボット上のカメラからメータの値を読み取るためには、画像からメータの領域を検出する必要がある。
本研究では、メータ領域検出技術として、形状ベース方式、テクスチャベース方式、背景情報ベース方式を開発し、形状・サイズとサイズの異なるメータに対してその有効性を比較検討した。
検討の結果、背景情報ベース方式がメータの形状と個数によらず最も遠くのメータを検出可能であり、メータの直径が40pxで撮影された状態でも安定して検出できることを確認した。
}
]


\section{はじめに}
\begin{figure}[htb]
 \centering
 \includegraphics[keepaspectratio, width=\linewidth]
      {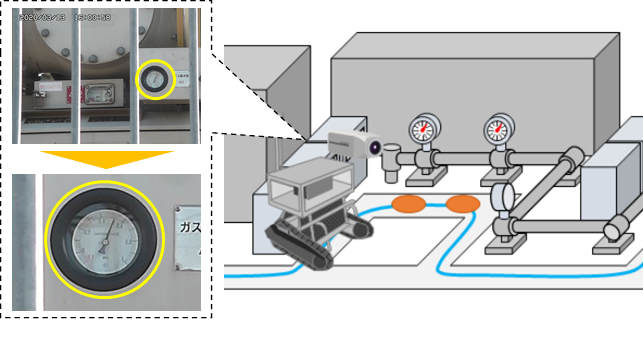}
 \caption{自律走行ロボットによるインフラ保守点検の様子}
 \label{fig:intro}
\end{figure}

日本において人口減少及び超高齢化が社会問題の一つとなっており、労働人口も今後加速度的に減少していくことが予想されている。社会インフラ設備の保守点検業務も例外ではなく、業務の効率化・自動化が急がれている。
社会インフラ点検を自動化する技術の一つとして、自律走行ロボットによる保守点検がある。
自動化が望まれる点検項目は複数存在するが、本研究ではその中で各種計器（メータ）の読み取りに着目する。
メータの読み取り手順は、図\ref{fig:intro}に示すように自律走行ロボットがインフラ施設を巡回し、予め指定したメータ撮影位置で静止した後にロボット上のPTZ(Pan, Tilt, Zoom)カメラを制御してメータの画像を取得し、メータ読み取りアルゴリズムを実行するというものである。
画像からメータ値を自動で読み取るアルゴリズム\cite{reading}の前提として、対象をある程度以上の大きさでカメラの視野内に捉える必要がある。
その実現手段としては、撮影地点ごとにPTZカメラの撮影方向とズーム倍率を登録しておくというものが考えられる。
しかし通常ロボットの自己位置推定には誤差が含まれることから静止位置にばらつきが生じ、前述の方法ではメータが画像範囲から外れてしまい値が読み取れないケースが発生しうる。
そこで本研究では図\ref{fig:intro}のように、まず対象メータの登録位置周辺を含む引きの撮影画像中からメータ領域を検出し、次のステップでメータ領域の中央にズームする方法を検討した。そのメータ領域の検出手法として3種類の候補を開発し、検出可能なメータの形状と画像中のサイズの観点で比較検討を行った。
\section{手法}
機械学習を用いたメータ検出手法における障害として、学習データの入手が困難であることが挙げられる。例えば変電所には多種多様なメータが存在するため、個別のメータに対して十分な量の画像データを入手することは難しいことが多い。また、実務的にはメータの画像を顧客先で取得しなければならない難しさも生じる。
そこで、少数のメータ画像からメータを検出できる手法を開発する。\par
本章では、開発した形状ベースの方式、テクスチャベースの方式、背景情報ベースの方式について、それぞれ説明する。
\subsection{形状ベース方式}
\begin{figure}[htb]
 \centering
 \includegraphics[keepaspectratio, width=\linewidth]
      {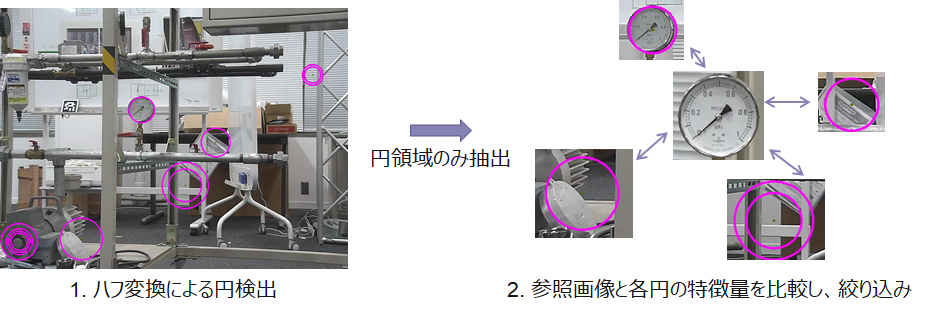}
 \caption{形状ベース手法}
 \label{fig:prop.hough.abs}
\end{figure}
形状ベースの手法は、ハフ変換\cite{duda1972use}により円検出を行った後、各円についてメータの参照画像と特徴量\cite{lowe1999object}を比較して絞り込むものである。図\ref{fig:prop.hough.abs}にその概要を示す。
\subsection{テクスチャベース方式}
\begin{figure}[htb]
 \centering
 \includegraphics[keepaspectratio, width=\linewidth]
      {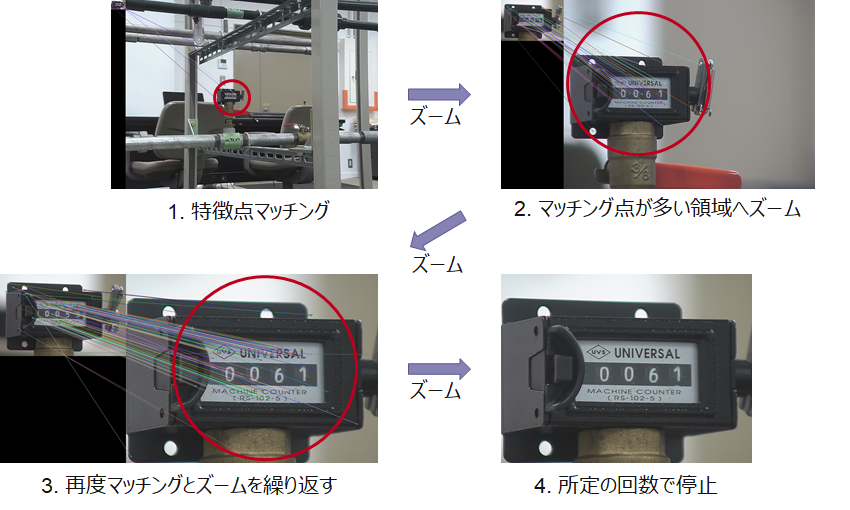}
 \caption{テクスチャベース方式}
 \label{fig:prop.filter.abs}
\end{figure}
検討するテクスチャベース方式はロボット・カメラの位置姿勢推定アルゴリズムである。メータ位置が記載された地図上におけるカメラの位置姿勢がわかれば、メータへの相対位置が定まり、メータを検出できる。
テクスチャベース方式の概要を図\ref{fig:prop.filter.abs}に示す。初めにロボットの自己位置推定アルゴリズム(SLAM等)から、ロボット・カメラの位置候補をいくつかそれぞれの有力度とともに保持しておく。そして初めの撮影画像上における参照画像とのマッチング点の位置と集中具合から、位置候補を絞り込み、有力度を更新する。次に最も有力な位置候補に従ってカメラを制御し、メータへある程度ズームする。ズーム後の撮影画像で再び特徴点マッチングを行い、マッチング点の位置と集中具合から位置候補をさらに絞り込み、更新していく。これをあらかじめ決めた回数繰り返すことでカメラの自己位置推定の精度を上げていき、最終的な推定位置の結果をもってメータ位置を検出する。
\subsection{背景情報ベース方式}
\begin{figure}[htb]
 \centering
 \includegraphics[keepaspectratio, width=\linewidth]
      {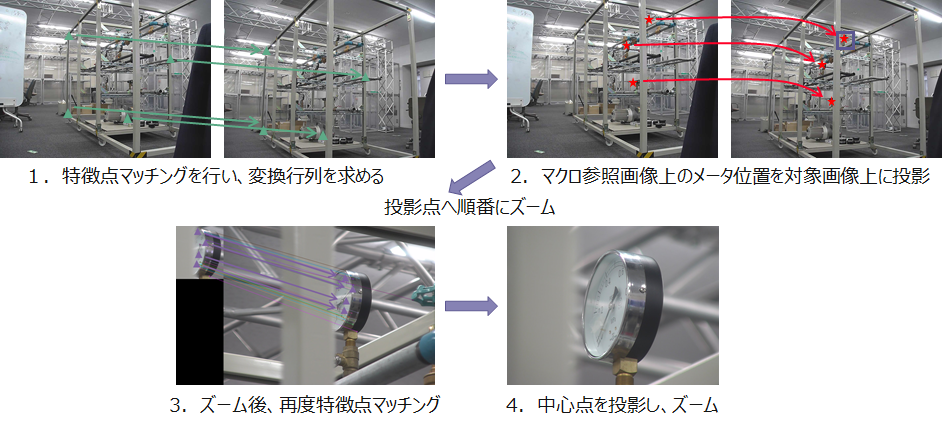}
 \caption{背景情報ベース方式}
 \label{fig:prop.macro.abs}
\end{figure}
背景情報ベース方式は画像上にメータが小さく写った場合にも対応した手法である。これを実現するため、背景情報を含むメータの周辺画像とその画像上にメータ位置をアノテーションしたものを追加で用意し、それを用いて大まかにズームした後、通常の特徴点マッチング\cite{lowe1999object}によりメータ検出位置を微調整するという方針をとる。図\ref{fig:prop.macro.abs}にその概要を示す。


\section{実験}
円形メータと矩形メータ対して、画像上のメータサイズを変えたときの３手法の検出率を計測した。円形メータは図\ref{fig:prop.hough.abs}、矩形メータは図\ref{fig:prop.filter.abs}で例示したものである。\ref{fig:exp.circle.result}に結果を示す。表中のハイフン(-)はそれより小さいメータで安定的に検出できたため、評価を省略したことを示している。
３方式の中では背景情報方式が最もメータサイズに対して頑健であり、メータの形状によらず直径が40px以下であっても安定して検出できていることがわかる。
\begin{table}[htb]
\begin{center}
\caption{円形メータに対する３方式の検出率}
  \begin{tabular}{|l||r|r|r|r|r|r|} \hline
    直径(pixel) & 160 & 120 & 100 & 80 & 60 & 40\\ \hline\hline
    形状 & 3/3 & 3/3 & 3/3 & 1/3 & 0/3 & 0/3 \\\hline
    テクスチャ & 3/3 & 2/3 & 2/3 & 1/3 & 0/3 & 0/3 \\\hline
    背景 & - & - & - & 3/3 & - & 3/3 \\\hline
  \end{tabular}\label{fig:exp.circle.result}
  \end{center}
\begin{center}
\caption{矩形メータに対する３方式の検出率}
  \begin{tabular}{|l||r|r|r|r|r|r|} \hline
    直径(pixel) & 160 & 120 & 100 & 80 & 60 & 40\\ \hline\hline
    形状 & 0/3 & - & - & - & - & - \\\hline
    テクスチャ & 3/3 & 3/3 & 0/3 & 0/3 & 0/3 & 0/3 \\\hline
    背景 & - & - & - & 3/3 & - & 3/3 \\\hline
  \end{tabular}\label{fig:exp.rectangle.result}
  \end{center}
\end{table}
\section{終わりに}
３方式の中では背景情報方式が最も小さいメータを検出できたが、背景画像にメータ位置を登録するコストがかかる。したがってメータサイズが120px以上のときは形状ベース手法やテクスチャベース手法を用いるといった運用上の組み合わせも考えられる。

\bibliographystyle{abbrv}
\bibliography{kenpo}
\simplefootnotetext{
(株)日立製作所　中央研究所\\
〒\hspace{0pt}185-8601 東京都国分寺市東恋ヶ窪1-280\\
E-mail:  yusuke.ohtsubo.nb@hitachi.com.jp}
\end{document}